\documentclass{article}
\usepackage{bbm}
\usepackage{amsmath,amssymb,amsfonts}
\usepackage{spconf,amsmath,graphicx}
\usepackage{booktabs}
\usepackage[T1]{fontenc}
\usepackage[ruled]{algorithm2e}
\usepackage{algpseudocode}  
\usepackage{dsfont}
\usepackage{marvosym}
\usepackage{float}
\usepackage{graphicx}
\usepackage{hyperref}
\usepackage{url}

\newcommand \pn{\emph{SR-init}}

\usepackage{threeparttable}

\usepackage{multicol}
\usepackage{multirow}
\usepackage[capitalize]{cleveref}

\usepackage{color}

\usepackage{verbatim}

\title{\pn{}: An Interpretable Layer Pruning Method}
\name{Hui Tang$^{\dagger}$ \qquad Yao Lu$^{\dagger}$~\thanks{$^\dagger$ Equal contribution.} \qquad Qi Xuan\textsuperscript{\Letter}}
  
\address{Institute of Cyberspace Security, Zhejiang University of Technology, Hangzhou, 310023. China}
\begin{document}
%
\maketitle
\begin{abstract}
Despite the popularization of deep neural networks (DNNs) in many fields, it is still challenging to deploy state-of-the-art models to resource-constrained devices due to high computational overhead. Model pruning provides a feasible solution to the aforementioned challenges. However, the interpretation of existing pruning criteria is always overlooked. To counter this issue, we propose a novel layer pruning method by exploring the \textbf{S}tochastic \textbf{R}e\textbf{-init}ialization. Our \textbf{\pn{}} method is inspired by the discovery that the accuracy drop due to stochastic re-initialization of layer parameters differs in various layers. On the basis of this observation, we come up with a layer pruning criterion, i.e., those layers that are not sensitive to stochastic re-initialization (low accuracy drop) produce less contribution to the model and could be pruned with acceptable loss. Afterward, we experimentally verify the interpretability of \pn{} via feature visualization. The visual explanation demonstrates that \pn{} is theoretically feasible, thus we compare it with state-of-the-art methods to further evaluate its practicability. As for ResNet56 on CIFAR-10 and CIFAR-100, \pn{} achieves a great reduction in parameters (63.98\% and 37.71\%) with an ignorable drop in top-1 accuracy (-0.56\% and 0.8\%). With ResNet50 on ImageNet, we achieve a 15.59\% FLOPs reduction by removing 39.29\% of the parameters, with only a drop of 0.6\% in top-1 accuracy. Our code is available at  \href{https://github.com/huitang-zjut/SR-init}{https://github.com/huitang-zjut/SR-init}.

\end{abstract}

\section{Introduction}
\label{sec:intro}
Despite DNNs having become the go-to technique in many fields including computer vision~\cite{maqueda2018event} and natural language processing~\cite{sutskever2014sequence}, their success comes at a price, as state-of-the-art models are primarily fueled by complicated models~\cite{DBLP:conf/naacl/DevlinCLT19,DBLP:conf/iclr/DosovitskiyB0WZ21}, which significantly limits the deployment on resource-constrained devices. Much effort has been made to design compact models to overcome the aforementioned challenge. Model pruning is one stream among them, which can be categorized into weight pruning~\cite{han2016eie,han2015learning}, filter pruning~\cite{liu2017learning,molchanov2019importance,huang2018data}, and layer pruning~\cite{chen2018shallowing,elkerdawy2020filter,xu2020layer}.

Weight pruning condenses over-parameterized models by dropping redundant weights, which makes weight matrices sparse without practical speedup on general-purpose hardware. Filter pruning seeks to prune redundant filters or channels, which is constrained by the original model's depth, as each layer must contain at least one filter or channel.
\begin{figure}[t]
  \centering
   \includegraphics[width=0.9\linewidth]{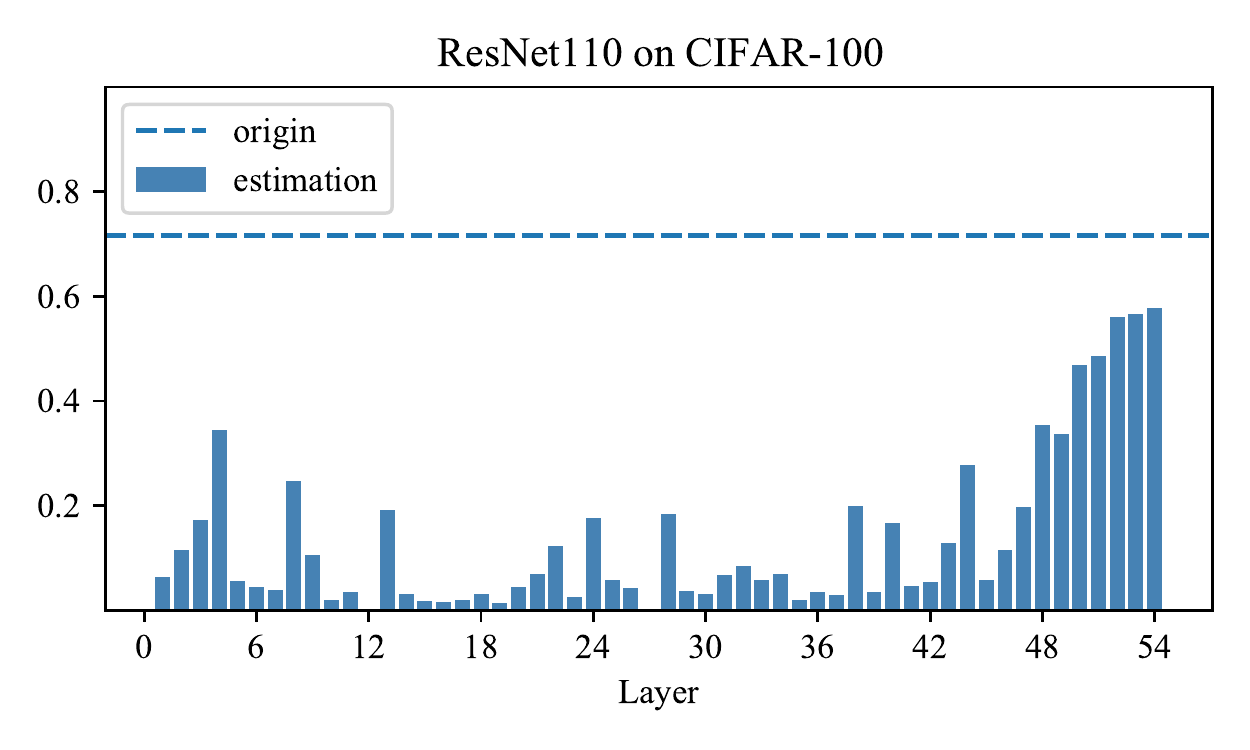}
  \setlength{\abovecaptionskip}{-2cm}
   \caption{Estimation accuracy with randomly re-initialized model on each layer. The dotted line represents the original accuracy for pre-trained ResNet110 on CIFAR-100.}
  \label{first} 
  \vspace{-0.5cm}  
\end{figure}
Compared to weight pruning and filter pruning, layer pruning removes entire redundant layers, which is more suitable for resource-constrained devices.
Hence, we concentrate on layer pruning for better inference speedup. Existing layer pruning methods differ in the selection of redundant layers. For example, Chen et al.~\cite{chen2018shallowing} estimate the layer importance using linear classifier probes~\cite{alain2016understanding}, prune the less important layers and recover model performance with knowledge distillation. Elkerdawy et al.~\cite{elkerdawy2020filter} utilize filter importance criteria instead to estimate a per-layer importance score in one-shot and fine-tune the shallower model.

These studies achieve empirical successes but are fundamentally limited, because they unconsciously ignore the interpretability of the proposed criteria. Actually, the most intuitive manifestation of layer redundancy embodies the duplicate stacking layers that perform the same operation repeatedly, refining features just a little bit each time. Embracing this viewpoint allows us to focus on how to estimate whether a layer performs the repeated operation. Intuitively speaking, if a layer only performs the repeated operation as the previous layer, then we believe that perturbing this layer will have little impact on model performance (especially for models with residual connections). Stochastic re-initialization of layer parameters can be regarded as an extreme form of perturbation, which arises our conjecture: \textbf{Can we characterize layer redundancy through the accuracy drop brought
by randomly re-initializing layer parameters?} To verify our conjecture, we perform a preliminary experiment by randomly re-initializing each layer of a pre-trained model. As shown in~\cref{first}, we observe that different layers exhibit accuracy drop at different magnitudes, which indicates that the sensitivity of different layers towards stochastic re-initialization varies (Further exploration will be comprehensively provided in~\cref{sec:Visual Interpretation}). Inspired by this, we propose a new layer pruning criterion, termed \pn{}, which seeks to prune layers that are not sensitive to stochastic re-initialization (low accuracy drop). Finally, extensive experiments demonstrate the feasibility of \pn{} for depicting layer redundancy.

To summarize, we make the following contributions: (1) We demonstrate the correlation between the accuracy drop induced by stochastic re-initialization and per-layer redundancy, and thus propose a simple criterion, i.e., \pn{}, for layer pruning; (2) With the guidance of \pn{}, we prune redundant layers in DNNs with a negligible performance loss and considerable model complexity reduction.


\section{METHODOLOGY}
\subsection{Preliminaries}
For the typical image classification task, the feedforward process of DNNs transfers the input sample into a classifier by sequentially mapping throughout consecutive layers. Therefore, we can model this process as follows: 
\begin{equation} \label{original model}
\begin{aligned}
\hat{y} = f_{n+1} \circ f_n \circ \cdots\circ f_1(x),
\end{aligned}
\end{equation}
where $\circ$ and $n$ denote function composition and the number of layers, respectively. $f_i: F_{i-1} \rightarrow F_i$ are maps between feature spaces $F_{i-1}$ and $F_i$ for $i \in [n]$ with $F_0 = x$, and $f_{n+1}$ is the classifier at the last layer. In this paper, the layer for ResNets refers to a bottleneck building block or a building block as Lu et al.\cite{lu2022understanding} in our work.
\subsection{\pn{} and Accuracy Drop}
For eliciting the final pruning criterion, we introduce some concepts in stochastic re-initialization. \pn{} should strictly obey the original distribution of layer parameters. Specifically, for a particular layer, the parameters get randomly re-initialized according to the activation and its corresponding distribution rule, while others are fixed. In this study, we utilize the Kaiming Initialization~\cite{he2015delving} to make sure the integrated layer parameters obey the normal distribution as \cref{Kaiming Initialize}.
\begin{equation} \label{Kaiming Initialize}
\begin{aligned}
f_i' \sim N\left(0, (\sqrt{\frac{2}{d_i}})^2\right),\\
\end{aligned}
\end{equation}
where $d_i$ represents the feature dimension of the corresponding layer. 
Through the stochastic re-initialization on the $i$-th layer, the new feedforward process can be constructed as:
\begin{equation} \label{re-initialization}
\begin{aligned}
     \hat{y^i} = f_{n+1} \circ f_n\cdots\circ f_i'\circ\cdots \circ f_1(x),
\end{aligned}
\end{equation}
where $\hat{y^i}$ represents the estimation model compared to the original model for $i$-th layer while other layers remain consistent and $f_i'$ represents the processed layer map by \pn{}.

After \pn{}, we could directly calculate the accuracy of estimation model just like the preliminary experiment in \cref{sec:intro}. However, merely utilizing the estimation accuracy can not reflect the changes caused by \pn{}. Therefore, we make a step further to evaluate the accuracy drop compared to the original model and regard it as $Drop =\{drop_1,\ldots,drop_i,\ldots,drop_n \}$. Specifically, for a given pre-trained DNN model with $n$ layers and dataset $D=\left\{\left(x_{j}, y_{j}\right)\right\}_{j=1}^{m}$, $\hat{y_p^i}(x_j) = argmax(\hat{y^i}(x_j))$ represents the forecast label according to the maximum probability value for input sample $x_j$. We can calculate the top-1 accuracy drop by \cref{D-Acc}.
\begin{equation} \label{D-Acc} 
\begin{aligned}
    drop_i &= \sum_{j=1}^{m}\frac{ \mathbb{I}\left(\hat{y_p}\left(x_{j}\right)=y_{j}\right) - \mathbb{I}\left(\hat{y^i_p}\left(x_{j}\right)=y_{j}\right)}{m}
\end{aligned}
\end{equation}
where $\mathbb{I}()$ refers to the indicator function that equals one when the equation inside holds and $drop_i$ denotes the top-1 accuracy drop between model $\hat{y^i}$ and $\hat{y}$. 

\subsection{\pn{} Pruning}
The pruning policy adopted in our work is to (1) prune the layer according to the \pn{} and threshold, and (2) stop pruning before the feature dimensions can not map aptly. The former is inferred that the layers with low accuracy drop embody more redundancy, which is oppositely less likely to compromise prediction ability if removed. The latter is motivated by the consideration that the model should have a consecutive feedforward process rather than a dimension mismatch.
\begin{algorithm}[t]
    \caption{\pn{} Pruning}
    \label{algorithm:layer selection}
    \LinesNumbered
    \KwIn {$\left\{f_i\right\}_{i=1}^{n+1}$: pre-trained layers and classifier}
    \KwIn {$d_i$: feature dimension of $i$-th layer}
    \KwIn {$t_{err}$: threshold for maximal accuracy drop}
    \KwIn {$\left\{\left(x_{j}, y_{j}\right)\right\}_{j=1}^{m}$: dataset}
    \KwOut {$pruned\_layer$}
    Initialize $pruned\_layer = \emptyset$, $cur = i = j = 1$\\    
    \While{$i <= n$}{ \label{li:loop_begin}
     $\hat{y} \leftarrow f_{n+1} \circ f_n \circ f_i\circ\cdots \circ f_1$\\
     $f_i' \leftarrow$ Re-init $\left(0,(\sqrt{\frac{2}{d_i}})^2\right)$\\
     $\hat{y^i} \leftarrow$ Re-bulid $\left(y, f_i', i\right)$\\
     \While{$j++ <= m$}{
     $\hat{y_p}(x_j) = argmax(\hat{y}(x_j))$\\
     $\hat{y_p^i}(x_j) = argmax(\hat{y^i}(x_j))$\\
     }
     $ drop_i = \sum_{j=1}^{m}\frac{ \mathbb{I}\left(\hat{y_p}\left(x_{j}\right)=y_{j}\right) - \mathbb{I}\left(\hat{y^i_P}\left(x_{j}\right)=y_{j}\right)}{m}$\\
     \If{$ drop_i < t_{err}$}{
      $pruned\_layer = pruned\_layer \cup cur$\\
     }
     $i++$\\
     $cur = i$\\
    } 
    \Return $pruned\_layer$
\end{algorithm}

With our policy in place, the procedure for layer pruning, detailed in \cref{algorithm:layer selection}, is straightforward and proceeds as follows. For the given DNN model $y=\left\{f_i\right\}_{i=1}^{n+1}$ with $n$ pre-trained layers and one classifier, dataset $D=\left\{\left(x_{j}, y_{j}\right)\right\}_{j=1}^{m}$ and feature dimension $d_i$, we randomly re-initialize each layer with Kaiming Initialization, where the $0$ and $(\sqrt{\frac{2}{d_i}}) ^2$ input to Re-init are the mean and variance of layer parameters' distribution. According to the constructed random layer $f_i'$, we could Re-build the original model $y$ at $i$-th layer. Before we calculate the accuracy drop by \cref{re-initialization} for each, $argmax()$ method reveals the most likely forecast label according to the maximum prediction value for input $\hat{y}(x_j)$ and $\hat{y^i}(x_j)$. Then, we keep adding redundant layers if the $drop_i$ is lower than the threshold value $t_{err}$, which is set artificially and indicates the maximal acceptance degree for accuracy drop. After receiving the $pruned\_layer$ and fine-tuning, we finally obtain the pruned model.

\section{Experiments}
In this section, we illustrate the feasibility of \pn{} for characterizing layer redundancy. Afterward, we evaluate the validity of this criterion in layer pruning with extensive experiments.
\subsection{Experimental Settings}
\textbf{Datasets.} We verify \pn{} with a family of ResNets~\cite{he2016deep} on CIFAR-10~\cite{krizhevsky2009learning}, CIFAR-100~\cite{krizhevsky2009learning}, and ImageNet~\cite{DBLP:journals/ijcv/RussakovskyDSKS15}. CIFAR-10 consists of 60K $32 \times 32$ RGB images (50K for training and 10K for testing) in 10 classes, while CIFAR-100 is 100 fine-grained classes with the same setting. ImageNet consists of $224 \times 224$ RGB images (1.2M for training and 50K for testing) in 1000 object classes.

\noindent \textbf{Implementation details.} In our experiments, we train ResNet56 on CIFAR-10 and CIFAR-100 using stochastic gradient descent (SGD)~\cite{bottou2012stochastic} algorithm with an initial learning rate of 0.01. The batch size, epoch, weight decay, and momentum are set to 256, 150, 0.005, and 0.9, respectively. For ImageNet, we use the pre-trained ResNet50 provided in torchvision~\footnote{https://github.com/pytorch/vision}. 

After layer pruning, we fine-tune all shallower models except for ResNet50 by using cosine annealing with warm restarts~\cite{DBLP:conf/iclr/LoshchilovH17}, where we use SGD for ResNet50 on ImageNet. As for performance evaluation, we adapt the number of parameters and required Float Points Operations (denoted as FLOPs) to calculate the model size and computational cost. The calculation of FLOPs and parameters is accomplished by a package in PyTorch~\cite{paszke2017automatic}, termed THOP~\footnote{https://github.com/Lyken17/pytorch-OpCounter}. 
\subsection{Visual Interpretation}
\label{sec:Visual Interpretation}
\begin{figure*}[t]
  \centering
   \includegraphics[width=0.99\linewidth]{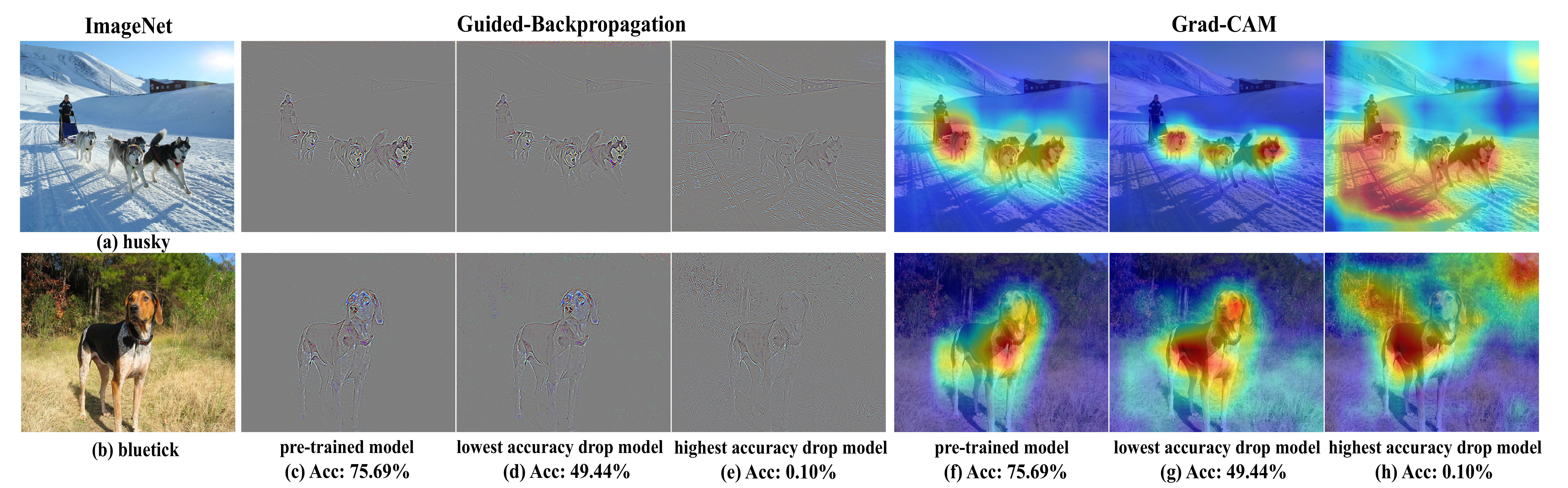}
   \caption{Visualization of Grad-CAM and Guided-Backpropagation for ResNet50 on ImageNet. (c, f): visualization of pre-trained model. (d, g): visualization of \pn{} on the lowest accuracy drop model. (e, h):  visualization of \pn{} on the highest accuracy drop model. The top-1 accuracy of different models is marked below the visual results.}
   \label{fig:visualization result}
\setlength{\belowcaptionskip}{-1cm}
\end{figure*}

This section is mainly to supplement the interpretation of the correlation between accuracy drop and layer redundancy. In order to make a reasonable explanation, we utilize two different fine-grained visualization methods in \cref{fig:visualization result}, Grad-CAM~\cite{selvaraju2017grad} and Guided-Backpropagation~\cite{springenberg2014striving}, respectively. The explanatory experiments are conducted on ImageNet with pre-trained ResNet50. We depict the visualization results for the most and least relevant layers determined by the \textbf{\pn{}} process. It turns out that the performance of Grad-CAM on the model with the lowest accuracy drop (26.25\%) still localizes class-discriminative regions correctly, and the Guided-Backpropagation’s maximal feature activation still matches with the input sample image, which means relative redundancy of this layer. Likewise, the opposite result appears at the highest accuracy drop (75.59\%) model, which gives rise to a great discriminative diffusion in Grad-CAM and reduction of the target activation in Guided-Backpropagation. Obviously, the model accuracy drop brought by the \pn{} is correlated to the redundancy of the model at layers.

\subsection{Main Results}
\label{sec:Main Results}
\begin{figure}[t]
  \centering
   \includegraphics[width=0.99\linewidth]{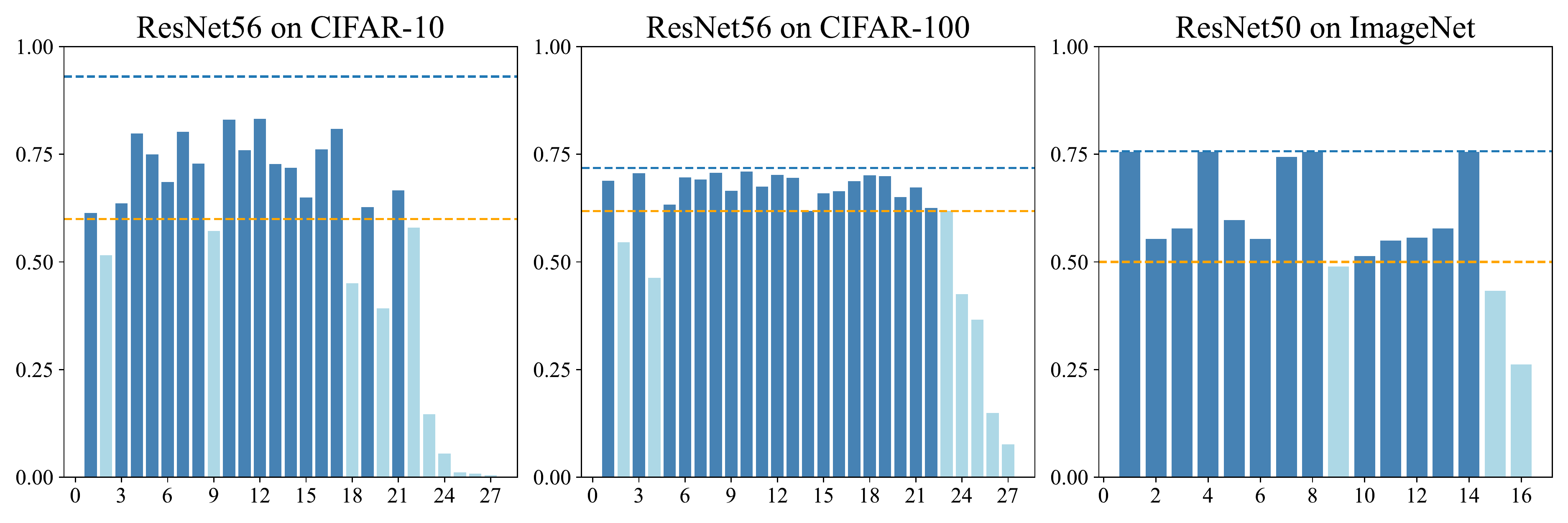}
   \caption{The accuracy drop of ResNet56 on CIFAR-10, CIFAR-100 and ResNet50 on ImageNet. The orange line embodies the maximal acceptance threshold. The light blue bar and dotted line represent layers below the threshold and original accuracy, respectively. }
  \label{layer_pruning} 
\vspace{-0.3cm} 
\end{figure}

To better understand the \pn{}, we illustrate this value in \cref{layer_pruning} calculated by \cref{D-Acc}. The result intuitively perceives the impact of \pn{} for prediction, and the dotted blue lines at the top represent the original accuracy for pre-trained ResNet model while orange lines show the empirically obtained value for maximal acceptance threshold. The layers with light color reveal that their accuracy drop is lower than the threshold, which means existing lower compromise on prediction ability and higher redundancy. Besides, redundant layers are generally gathered at the end of the network. The main reason is that the ResNet block tends to perform iterative refinement of features at deeper layers, which are more likely to arise layer redundancy and parameter stacking~\cite{DBLP:conf/iclr/JastrzebskiABVC18}. This property makes our method more advantageous for the parameter reduction since the numerous parameters are concentrated in deep layers. 

Compared with other recent state-of-art pruning works, the results of the \pn{} method for ResNet56 on CIFAR-10, CIFAR-100 and ResNet50 on ImageNet are shown in~\cref{cifar10-100} and~\cref{imagenet}, respectively. For all cases, we observe that our method surpasses Chen et al~\cite{chen2018shallowing} and Xu et al~\cite{xu2020layer} on CIFAR-10 and Elkerdawy et al~\cite{elkerdawy2020one} on ImageNet in all aspects. Compared to DBP~\cite{DBLP:journals/corr/abs-1912-10178} and Lu et al~\cite{lu2022understanding}, our method still obtains an excellent top-1 accuracy (93.83\% vs. 93.27\% and 93.38\%), which is even better than the baseline model (93.83\% vs. 93.27\%). Besides, although our method is inferior to DBP~\cite{DBLP:journals/corr/abs-1912-10178} and Lu et al.\cite{lu2022understanding} in FLOPs reduction, we take an obvious lead in parameters reduction (63.98\% vs. 40.00\% and 43.00\%). Similar observations can be found in the experiment on CIFAR-100, \pn{} leads to an improvement over the method proposed by Chen et al~\cite{chen2018shallowing} in accuracy (71.01\% vs. 69.78\%) and parameter reduction (37.71\% vs. 36.10\%). However, different pruning methods have pros and cons, and there is no absolute winner in all settings. \pn{} achieves the optimal parameter reduction in all methods with an acceptable accuracy loss. Finally, considering the results on ImageNet, our method exhibits competitive or even better performance with other state-of-art models. In comparison with Xu et al.\cite{xu2020layer}, our method advances in better parameters and FLOPs reduction while maintaining a similar accuracy (75.41\% vs. 75.44\%). In general, extensive experiments on various models demonstrate the effectiveness of \pn{} in reducing computational complexity and model size.
\vspace{-3mm}

\begin{table}[t] 
\vspace{-6mm}
	\caption{Pruning results of ResNet56 on CIFAR-10 and CIFAR-100. PR is the pruning rate.}
	\scalebox{0.9}{
		\centering 
		\begin{tabular}{lccc}
			\toprule
			\multicolumn{1}{l|}{Algorithm}  & Acc(\%) &  Params(PR)$\%$  & FLOPs(PR)$\%$ \\
			\midrule
			\multicolumn{4}{c}{CIFAR-10} \\
			\midrule
			\multicolumn{1}{l|}{ResNet56}  & 93.27 & 0 & 0 \\
			\multicolumn{1}{l|}{DBP~\cite{DBLP:journals/corr/abs-1912-10178}}  & 93.27 & 40.00 & 52.00 \\
			\multicolumn{1}{l|}{Chen et al~\cite{chen2018shallowing}}  & 93.29 & 42.30 & 34.80 \\
			\multicolumn{1}{l|}{Xu et al~\cite{xu2020layer}} & 93.75 & 28.20 & 36.00 \\
			\multicolumn{1}{l|}{Lu et al~\cite{lu2022understanding}}  & 93.38 & 43.00 & \textbf{60.30} \\
			\multicolumn{1}{l|}{Ours} & \textbf{93.83} & \textbf{63.98} & 37.51 \\
			\midrule
			\multicolumn{4}{c}{CIFAR-100} \\
			\midrule
			\multicolumn{1}{l|}{ResNet56} & \textbf{71.81} & 0 & 0 \\
			\multicolumn{1}{l|}{Chen et al~\cite{chen2018shallowing}} & 69.78 &36.10 &\textbf{38.30} \\
			\multicolumn{1}{l|}{Lu et al~\cite{lu2022understanding}}  & 71.39 & 9.20 & 30.16\\
			\multicolumn{1}{l|}{Ours}  & 71.01 & \textbf{37.71} & 26.28 \\
		
			\bottomrule
		\end{tabular}
	}
	\label{cifar10-100}
\vspace{-6mm}
\end{table}

\begin{table}[t]
\caption{Pruning result of ResNet50 on ImageNet.}
	\scalebox{0.85}{
		\centering 
		\begin{tabular}{lccc}
			\toprule
			\multicolumn{1}{l|}{Algorithm}  & Acc(\%) &  Params(PR)$\%$ & FLOPs(PR)$\%$ \\
			\midrule
			\multicolumn{1}{l|}{ResNet50}  &\textbf{76.01}  &  0 & 0 \\
						\multicolumn{1}{l|}{Elkerdawy et al~\cite{elkerdawy2020one}} & 74.74 & 8.24 & -\\
						\multicolumn{1}{l|}{Xu et al~\cite{xu2020layer}} & 75.44 & 35.02 & 11.19\\
			\multicolumn{1}{l|}{Ours} & 75.41 &\textbf{39.29}  &\textbf{15.59} \\ 
		
			\bottomrule
		\end{tabular}
	}
	\label{imagenet}
\vspace{-5mm}
\end{table}

\section{CONCLUSION}
\vspace{-3mm}
In this work, via discrepant accuracy drop caused by \pn{}, we provide a new perspective to better understand layer redundancy with reasonable interpretation. Through extensive experiments, we demonstrate the feasibility and practicality of \pn{} for layer pruning. The end goal of this work is to facilitate more researchers to concentrate on model optimization in an interpretable manner.

\section{ACKNOWLEDGEMENT}
\vspace{-3mm}
This work was supported in part by the Key R\&D Program of Zhejiang under Grant 2022C01018, and by the National Natural Science Foundation of China under Grants U21B2001 and 61973273.


\bibliographystyle{IEEEbib}
\bibliography{IEEE}

\end{document}